\documentclass[letterpaper, 10 pt, conference]{ieeeconf}  

\usepackage{graphics} 
\usepackage{epsfig} 
\usepackage{mathptmx} 
\usepackage{times} 
\usepackage{amsmath} 
\usepackage{amssymb}  
\usepackage{graphicx}
\usepackage{siunitx}
\usepackage{dblfloatfix}
\usepackage{multirow}
\usepackage{multicol}
\usepackage{xspace} 
\usepackage{booktabs}
\usepackage{longtable}
\usepackage{listings}
\usepackage{authblk}
\usepackage{comment}
\usepackage{epstopdf}

\usepackage[caption=false]{subfig}

\IEEEoverridecommandlockouts                              

\title{\LARGE \bf
AR Training App for Energy Optimal Programming of Cobots
}

\author{Juan Heredia$^{1}$, Christian Schlette$^{1}$ and Mikkel Baun Kjærgaard$^{1}$
\thanks{$^{1}$Authors are with the Mærsk Mc-Kinney Møller Institure,
        University of Southern Denmark, Odense, Denmark
        {\tt\small jehm@mmmi.sdu.dk, chsch@mmmi.sdu.dk, mbkj@mmmi.sdu.dk}}
\thanks{
        This extended abstract contribution has been accepted to the 1st International Workshop “Horizons of an Extended Robotics Reality” held at IEEE/RSJ IROS 2022, Kyoto, Japan.
        }}

\begin{document}

\maketitle
\thispagestyle{empty}
\pagestyle{empty}

\begin{abstract}

Worldwide most factories aim for low-cost and fast production ignoring resources and energy consumption. But, high revenues have been accompanied by environmental degradation. The United Nations reacted to the ecological problem and proposed the Sustainable Development Goals, and one of them is Sustainable Production (Goal 12). In addition, the participation of lightweight robots, such as collaborative robots, in modern industrial production is increasing. The energy consumption of a single collaborative robot is not significant, however, the consumption of more and more cobots worldwide is representative. Consequently, our research focuses on strategies to reduce the energy consumption of lightweight robots aiming for sustainable production. Firstly, the energy consumption of the lightweight robot UR10e is assessed by a set of experiments. We analyzed the results of the experiments to describe the relationship between energy consumption and the evaluation parameters, thus paving the way for optimization strategies. Next,  we propose four strategies to reduce energy consumption: 1) optimal standby position, 2) optimal robot instruction, 3) optimal motion time, and 4) reduction of dissipative energy. The results show that cobots potentially reduce from 3\% up to 37\% of their energy consumption, depending on the optimization technique. To disseminate the results of our research, we developed an AR game in which the users learn how to program cobots energy-efficiently.

\end{abstract}

\section{INTRODUCTION}Traditionally, industrial production has aimed to fast production and low production costs. But, this production came with environmental degradation. Governments have reacted by creating policies to reduce resources and energy consumption (EC) in production facilities worldwide. For example, the United Nations proposed sustainable development goals (SDG), which offer guidelines to reach sustainable production (Goal 12). Currently, factories need to restructure their processes and use more efficient machines or efficient activity scheduling to reduce energy and resource consumption. Even new technologies like collaborative robots are inefficient and can be modified to reduce their EC.

\begin{figure}[thb]
\centering
\includegraphics[width=0.48 \textwidth]{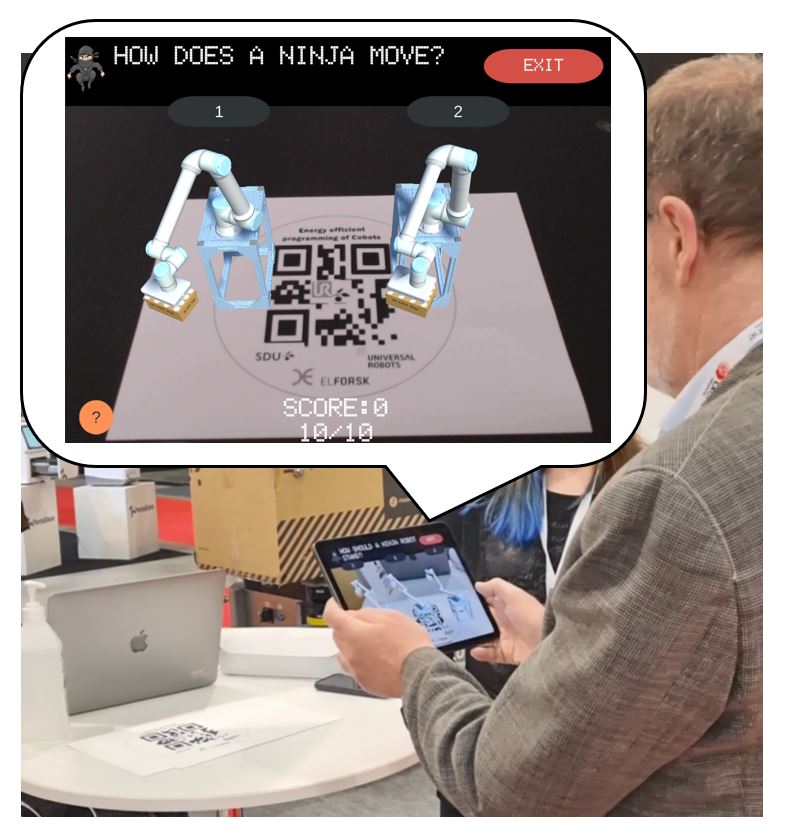}
\caption{Training of Practitioners for Energy Optimal Programming.}
\label{fig:Practitioners}       
\end{figure}

In recent years, the energy optimization of traditional industrial robots has been the focus of many authors. We hypothesize that some techniques applied to conventional industrial robots might be extrapolated to lightweight robots, but they should be tested to prove their feasibility. Lightweight robots and conventional industrial robots have many similarities, but their constructive differences (type of motors, link weights, payload capacity, materials) cause each robot's EC to differ. A giant and weighty robot's dynamic behavior certainly is different from a small light robot. Gaspareto et al. said that the inertial and elastic terms are usually neglected in a heavy rigid link configuration.\cite{gaspareto} 

We have focused our attention on lightweight robots because they will play a vital role in the industry’s future. The EC of a single LIR is not significant, but worldwide thousands of LIRs are used in factories. Besides, LIRs are involved in applications, such as mobile manipulators, with a limited energy source. Energy-optimal behavior is required to extend the robot’s working time.

Firstly, we assessed the EC of the lightweight robot UR10e using the following evaluation criteria: joint configuration, payload, movement command, acceleration limit, velocity limit, and trajectory planning. Using the assessment results, we propose four techniques to reduce the EC: 
\begin{itemize}
    \item Optimal manufacturer-defined commands: We found out that it is recommendable to use joint linear movements instead of Cartesian linear movements, especially for fast moves. In the results of the experiments, this method saves up to 37\% (without payload) and 26\% (with payload) of EC. 
    \item Optimal motion time: In this method, we proposed to obtain the characteristic curve (EC vs. execution time) of a given path. Then, it is possible to get the optimal scale factor that minimizes the robot EC. The results show that the robot saves up to 3.29\% (without payload) and 2.77\% (with payload) of the maximum experimental energy consumption.
    \item Reduction of dissipating energy: Robots can regenerate electrical energy when they are breaking, or going down (reduction of altitude). But, many robots can not send this energy back to the grid, and this energy is transformed into heat using a resistor. In this technique, the dissipative energy is avoided. The dynamic power saturator transforms all the regenerated energy (potential energy or kinematic breaking energy) into kinematic energy. The proposed control scheme reduces the energy consumption by 5.27\% in the proposed experiment. Besides, the temperature of the cabinet is reduced. Potentially, the energy for ventilation might be reduced. 
    \item Optimal standby position: The energy consumption of a robot in a standby position depends on the gravitational torque and the motor characteristics. Therefore, the joint configuration, which minimizes the gravitational torque, potentially consumes less energy. The experiments showed that it is possible to save up to 16\% of energy.
\end{itemize}

We claim that Augmented Reality is the right tool to train on how to apply the developed techniques due to three principal reasons: 1) Portability: it is not required to have a real robot. 2) Trainee can compare the performance of many robots in the same scene. 3) Engagement: users are engaged in using new technologies. \cite{AR_benefits} Therefore, we decided to use Augmented Reality for developing a training tool for Energy Optimal Programming. 

\begin{figure}[thb]
\centering
\includegraphics[width=0.5 \textwidth]{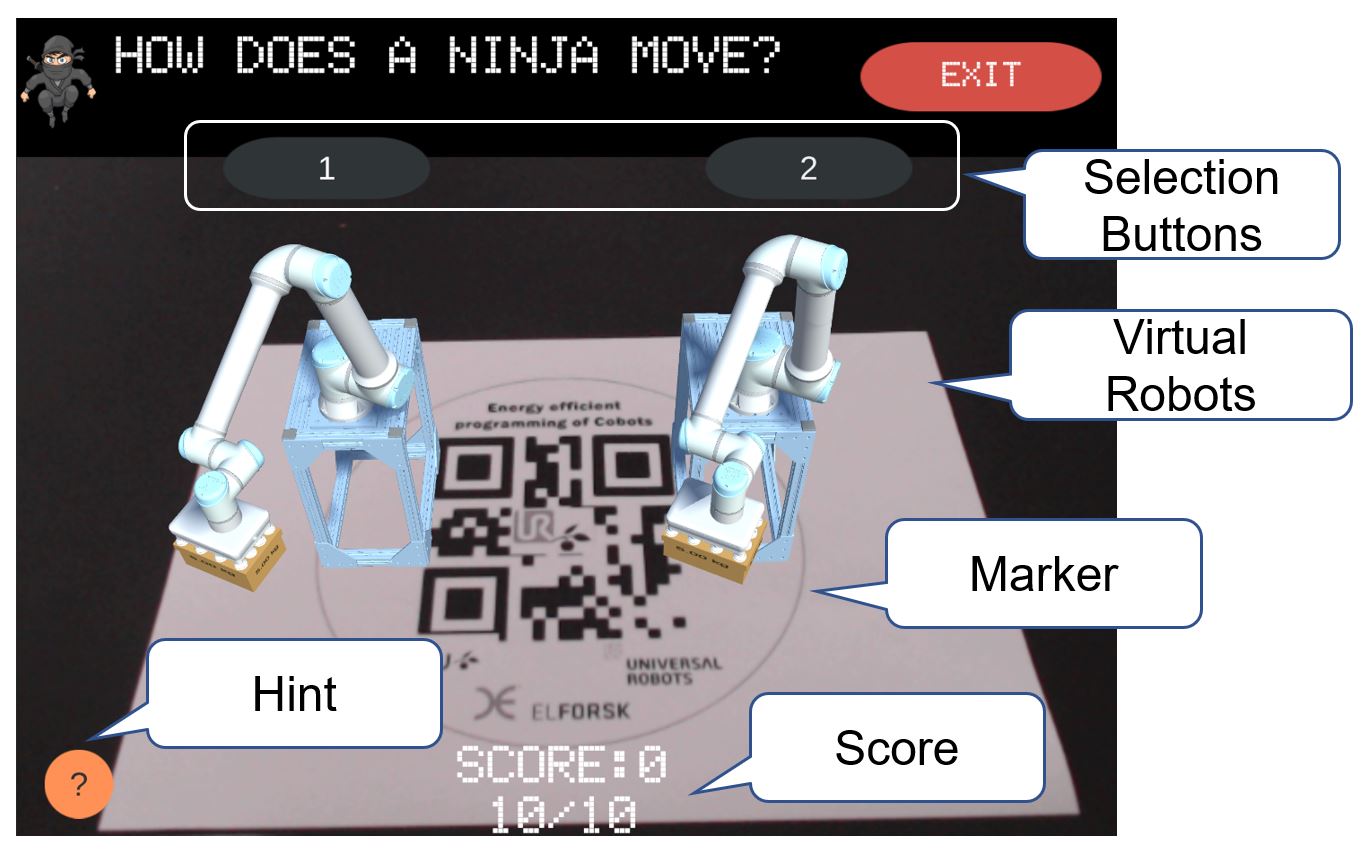}
\caption{Example scene of AR app.}
\label{fig:Escene}       
\end{figure}

The app was designed using Unity Software for IOs devices. The app was designed based on three principal components: 
\begin{itemize}
    \item Tracking: Vuforia library was employed to track a marker, which is shown in Fig. \ref{fig:Escene}.
    \item Virtual 3D Models: The 3D models of the robots were obtained from the official manufacturer's website. Models of UR3e and UR10e were utilized in the app.
    \item Database: The virtual robots access real operation data. We recorded data from real robot operations. In the dataset, there are joint positions, velocities, currents, motor temperatures, and energy consumption. 
\end{itemize}

The application is composed of ten scenes (see Fig. \ref{fig:Escene}), which present examples of the optimization techniques. In each scene, the user selects which robot consumes less energy from two or three options. One of the options applies the technique or techniques that reduce the robot's energy consumption. Additionally, users have access to the technique's theoretical foundation. 

Using AR, the trainee can observe information from the robot's performance that is not visible in the real robot. For example, the robot path was highlighted using an imaginary 3D line. If the training would be carried out using real robots, the trainee would have to imagine this 3D line. The use of AR allows the creation of visual features that facilitate the learning process.

In the fair R22 \cite{odense}, this application was presented to technical practitioners, as Fig.1 shows. We expect technicians to include the game's knowledge in their future Cobot applications. At the fair, 26 practitioners played the game. The average success rate was 76.8 \% and the average time was 6 minutes and 53 seconds. Most of the practitioners were experts or work with robots. We hypothesize that the success rate is high due to the robotics knowledge of the participants.

\section{CONCLUSIONS}

The use of AR in training and teaching is beneficial. It is not required to have a physical real robot. Also, the trainee can compare many robots doing different activities in the same scenario. It would be costly to have real robots doing the same task.

AR allows making abstract attributes real. The use of imaginary lines or attributes is a visual feature that enhances the trainee's learning process. This capability can be used in other applications such as robot troubleshooting, robot commissioning, and others.   

The application was used only for training. However, the application can be used to simulate and test new trajectories and paths looking for the optimal one. The application needs to include an energy consumption model of the robot. Then, the application can be used not only as a visualization tool.


\typeout{}
\bibliographystyle{IEEEtran}
\bibliography{mybib.bib}{}

\end{document}